\UseRawInputEncoding
\documentclass[10pt,twocolumn,letterpaper]{article}

\usepackage{cvpr}
\usepackage{times}
\usepackage{epsfig}
\usepackage{graphicx}
\usepackage{amsmath}
\usepackage{amssymb}
\usepackage{cite}
\usepackage{caption}
\usepackage{subfigure}
\usepackage{booktabs}
\usepackage{multirow}
\usepackage{diagbox} % 加载宏包

\renewenvironment{thebibliography}[1]{%
   \begin{oldthebibliography}{#1}%
     \setlength{\itemsep}{-.8ex}%
}%
{%
   \end{oldthebibliography}%
}

\usepackage[breaklinks=true,bookmarks=false]{hyperref}

\cvprfinalcopy % *** Uncomment this line for the final submission

\def\cvprPaperID{****} % *** Enter the CVPR Paper ID here

\begin{document}

%%%%%%%%% TITLE
\title{LNSMM: Eye Gaze Estimation With Local Network Share Multiview Multitask}

\author{Yong Huang, Ben Chen, Daiming Qu\\
Department of Electronics and Information Engineering,\\
HuaZhong University of Science and Technology\\
{\tt\small $\big\{huangyong,benchen,qudaiming\big\}$@hust.edu.cn}
}

%\and

\maketitle
%\thispagestyle{empty}

%%%%%%%%% ABSTRACT
\begin{abstract}
   Eye gaze estimation has become increasingly significant in computer vision.In this paper,we systematically study the mainstream of eye gaze estimation methods,propose a novel methodology to estimate eye gaze points and eye gaze directions simultaneously.First,we construct a local sharing network for feature extraction of gaze points and gaze directions estimation,which can reduce network computational parameters and converge quickly;Second,we propose a Multiview Multitask Learning (MTL) framework,for gaze directions,a coplanar constraint is proposed for the left and right eyes,for gaze points,three views data input indirectly introduces eye position information,a cross-view pooling module is designed, propose joint loss which handle both gaze points and gaze directions estimation.Eventually,we collect a dataset to use of gaze points,which have three views to exist public dataset.The experiment show our method is state-of-the-art the current mainstream methods on two indicators of gaze points and gaze directions.
\end{abstract}
%%%%%%%%% BODY TEXT
\section{Introduction}
Scientific experiments have shown that the human eye is in constant
motion at all times, which ensures that we can dynamically obtain the
information of the eye's line of sight in real time.However, this
information is affected by factors such as location and environment[1],
and the brain is processing this complex information At times, different
signals are generated, and these complex signals will cause our brains to
produce different stimuli, leading to changes in our psychological
conditions and different behaviors. The purpose of the human eye gaze
estimation technology is to analyze human behavior and predict the
behavior that will occur based on the characteristics of the eye. The
research and experiment on the characteristics of eye movement began in
the early part of the last century. The research on the related concepts
of eye gaze estimation began in the 1930s. At that time, due to the
backwardness of related technology and equipment, the research methods
were mainly limited to recording eye movement Analysis of trajectory and
movement-related data, only a few simple experiments can not get accurate
results. \par
\begin{figure}[h]   % 必须要有[h]否则插入的图片都在文字前面
    \centering  % 图像居中
    \includegraphics[width=4cm]{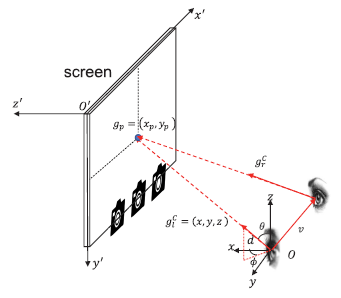}   %[]里可以指定影像大小
    \caption{Gaze point position depends on the gaze direction and eye position.
O is the eye position in the 3-D coordinate system}    % 图名
    \label{fig_electronic}  % 用于内部引用的图名
\end{figure}
The human eye gaze estimation technology is mainly used in human-computer
interaction and intelligent diagnosis analysis [5]. Common areas of human-computer interaction, for example, by capturing and estimating the gaze point of the human eye, it is possible to use eyeball rotation instead of traditional mouse clicks to operate the computer. The gaze point is equivalent to the mouse pointer of the computer. When we stare at a certain point, the pointer is fast Move and point to the point; it is also widely used in the medical field, such as wearing a prosthesis for the disabled, using the eye gaze estimation technology to capture the person’s gaze point and then feedback the corresponding coordinate position to the prosthesis, and the prosthesis receives the coordinate information to send Related commands are issued to complete the corresponding actions, etc.; Intelligent diagnosis analysis is currently used in smart traffic, and the driver's attention is detected in smart traffic. For example, by tracking and analyzing the driver’s eye movement, estimating its gaze direction and gaze point, it can be judged in real time whether the driver’s attention is concentrated, and by analyzing the eye movement data during driving, it can then be judged whether the driver is Illegal driving such as drunk driving and fatigue driving. For conventional inattention, we can send out corresponding prompts in real time to remind the driver to concentrate. For drunk driving or fatigue driving, we can issue a red warning and limit the speed of the car. In addition, real-time dynamic eye tracking can also avoid a series of wrong operations, such as the driver's steering wheel direction wrong due to factors such as tension, and track the driver's line of sight through real-time and accurate eye movements to capture his gaze direction and gaze Points can be forced to correct incorrect operations in critical situations to avoid driving accidents.

The estimation of the gaze point requires the determination of the pose of the head and the geometric calibration of the gaze direction. However, the existing algorithms either only consider the estimation of the gaze point or only the estimation of the gaze direction, but ignore the important relationship between the two. Based on the above discussion, this article aims to explore the correlation between the human eye's 2D gaze point and 3D gaze direction estimation, and build a human eye gaze estimation algorithm based on convolutional neural network to provide an innovative solution for obtaining accurate human eye gaze tracking. Program. This is for the research of psychology and neuroscience research, human-computer interaction, virtual reality and augmented reality, develop and improve the design theory of human eye gaze point estimation, promote the development of human-computer interaction, intelligent diagnosis and analysis, and realize the real use of machines by humans. The purpose of arbitrary control is important.

\section{Related work}

% All text must be in a two-column format. The total allowable width of the
% text area is $6\frac78$ inches (17.5 cm) wide by $8\frac78$ inches (22.54
% cm) high. Columns are to be $3\frac14$ inches (8.25 cm) wide, with a
% $\frac{5}{16}$ inch (0.8 cm) space between them. The main title (on the
% first page) should begin 1.0 inch (2.54 cm) from the top edge of the
% page. The second and following pages should begin 1.0 inch (2.54 cm) from
% the top edge. On all pages, the bottom margin should be 1-1/8 inches (2.86
% cm) from the bottom edge of the page for $8.5 \times 11$-inch paper; for A4
% paper, approximately 1-5/8 inches (4.13 cm) from the bottom edge of the
% page.
Early human gaze direction estimation mainly used physical methods to build models for measurement. Such methods are susceptible to external environmental influences and human interference, which often leads to low measurement accuracy. With the rapid development of electronic technology and computer technology, eye movement measurement methods and technical levels are constantly improving, especially with the rapid development of computer vision, more accurate and generalized measurement methods are becoming more and more mature. The gaze estimation methods are mainly divided into: feature-based gaze estimation technology, natural light-based gaze estimation technology, appearance-based gaze estimation technology. Feature-based gaze estimation technology can be further subdivided into: gaze estimation method based on two-dimensional regression, gaze estimation method based on three-dimensional modeling[16]
%-------------------------------------------------------------------------
\subsection{Feature-based gaze estimation method}
The gaze method that uses local features such as contours and corners of the eyes to be extracted from the eye image is called the feature-based method. The advantage of this method is that it can easily obtain the local features of the iris and pupil, as well as the geometric shape of the eye and some related eye physiological information , And these characteristics directly affect the effect of gaze estimation. Due to the high accuracy and practicality of this method, it has become the most researched method. Feature-based methods can be divided into gaze estimation based on geometric models and gaze estimation based on interpolation regression. Assuming that the mapping from image features to gaze coordinates (2D or 3D) has specific parameter forms, such as polynomials or neural networks, these methods avoid explicit calculation of the intersection between the gaze direction and the stared object. The method based on the three-dimensional model is based on the geometric model of the eye, which directly calculates the gaze direction from the characteristics of the eye, and the gaze point is estimated by the intersection of the gaze direction and the viewed object.

% \begin{verbatim}
% %\ifcvprfinal\pagestyle{empty}\fi
% \setcounter{page}{4321}
% \end{verbatim}
% where the number 4321 is your assigned starting page.

% Make sure the first page is numbered by commenting out the first page being
% empty on line 46
% \begin{verbatim}
% %\thispagestyle{empty}
% \end{verbatim}

%-------------------------------------------------------------------------
\subsection{Gaze estimation method based on natural light}
The natural light method uses infrared light source IR for natural replacement. The natural light method faces some new challenges, such as changes in the visible light spectrum and low-contrast images, but it is less sensitive to infrared light in the environment, so it may be more suitable for outdoor use. Colombo et al. [27] modeled the visible part of the user's eyeball as a plane, and regarded any eye movement caused by eyeball rotation as a translation of the pupil on the plane. Knowing the existence of a one-to-one mapping between the hemisphere and the projective plane, Pece [27] maps the gaze point as the viewpoint from the center of the iris to the monitor. This is only an approximation, because nonlinear one-to-one mapping is not considered. These methods are not invariants of head posture. Newman et al. [28] proposed two independent systems that use stereo and face models to estimate the gaze direction. The eyeball is regarded as a sphere, and the intersection of the gaze directions of the two eyes is the estimated gaze point. The eyeball center is estimated by a head pose model, and personal calibration is also used. Wang [29] and others also combined the face pose estimation system with a narrow field of view camera, assuming that the scleral contour is a circle, and using a new eyeball model to estimate its gaze direction in three-dimensional space.

\subsection{Appearance-based gaze estimation method}
%\noindent 段首前取消空格；\textbf 表示加粗
Feature-based methods need to detect pupils and flicker, but the extracted features may be error-prone. In addition, there may be some potential features that have not been extracted, but they play an important role in the transmission of gaze information. The accuracy of methods based on natural light is often limited, so there are more methods based on appearance. Similar to the appearance model of the eyes, the appearance-based gaze estimation model uses image content as input, with the purpose of directly mapping these content to screen coordinates. Therefore, we hope to be able to implicitly extract potential functions for estimating gaze points, related features, and personal changes without the need for scene geometry and camera calibration. One method is to use cropped eye images to train regression functions, such as multi-layer networks or Gaussian processing or popular learning. An image is a high-dimensional representation of data, which is defined on a low-dimensional manifold. Researchers use local linear embedding to learn the manifold of the human eye image. The number of calibration points they use is significantly reduced while improving accuracy. Williams et al. [26] used a sparse Gaussian process interpolation method to filter the visible spectrum image to obtain the gaze prediction value and related measurement errors.
Appearance-based methods generally do not require calibration of the camera and geometric data, because the mapping is performed directly on the image content, similar to interpolation-based methods. The appearance model must deduce the geometric shape and related features from the image, so a large number of calibration points are often required. Although the appearance-based method can establish a geometric model well, it cannot solve the problem of invariance of head posture. The reason is that in different postures and gaze directions, the appearance of the eye area looks the same. In addition, under the same posture, changes in lighting will also change the appearance of the eyes, which may lead to a decrease in accuracy.
The appearance-based gaze estimation model uses the image content as input, and the convolutional neural network can better realize the feature extraction of the eye image. Therefore, the current gaze estimation based on deep learning is mostly the appearance-based gaze estimation model.

%-------------------------------------------------------------------------
% \begin{table}
% \begin{center}
% \begin{tabular}{|l|c|}
% \hline
% Method & Frobnability \\
% \hline\hline
% Theirs & Frumpy \\
% Yours & Frobbly \\
% Ours & Makes one's heart Frob\\
% \hline
% \end{tabular}
% \end{center}
% \caption{Results.   Ours is better.}
% \end{table}

%-------------------------------------------------------------------------
% \subsection{Illustrations, graphs, and photographs}
% When placing figures in \LaTeX, it's almost always best to use
% \verb+\includegraphics+, and to specify the  figure width as a multiple of
% the line width as in the example below
% {\small\begin{verbatim}
%   \usepackage[dvips]{graphicx} ...
%   \includegraphics[width=0.8\linewidth]
%                   {myfile.eps}
% \end{verbatim}
% }
%------------------------------------------------------------------------
\section{Method}
\subsection{Architecture}
The global network of multi-view and multi-task human eye gaze points based on deep convolutional neural network includes the shared local network ResNext-36, which  are 36 convolutional layers and 3 down-sampling convolutional layers.The shared local network is used to extract the features of the gaze point and gaze direction datasets respectively, and then load the dataset features of the gaze point and the dataset feature of the gaze direction into the gaze point estimation network model and the gaze direction estimation network model, respectively, and stare Point estimation network and gaze direction estimation network use convolutional neural network for supervised learning, and stop training when the network model reaches saturation.
Gaze estimation network consist of shared local network,gaze direction estimation 
network and gaze point estimation network.As show in Fig. 2,we propose to feed both gaze direction data set and gaze point data set into shared local network ResneXt-36, make use of MTL framework to achieve propagation.Specifically, the extracted features by local shared CNN feed into two network,which can divided into two modules based on their function:gaze direction and gaze point estimation modules in Fig. 3.Next,we would detail three modules respectively.
\begin{figure*}[h]   % 必须要有[h]否则插入的图片都在文字前面
    \centering  % 图像居中
    \includegraphics[width=16cm]{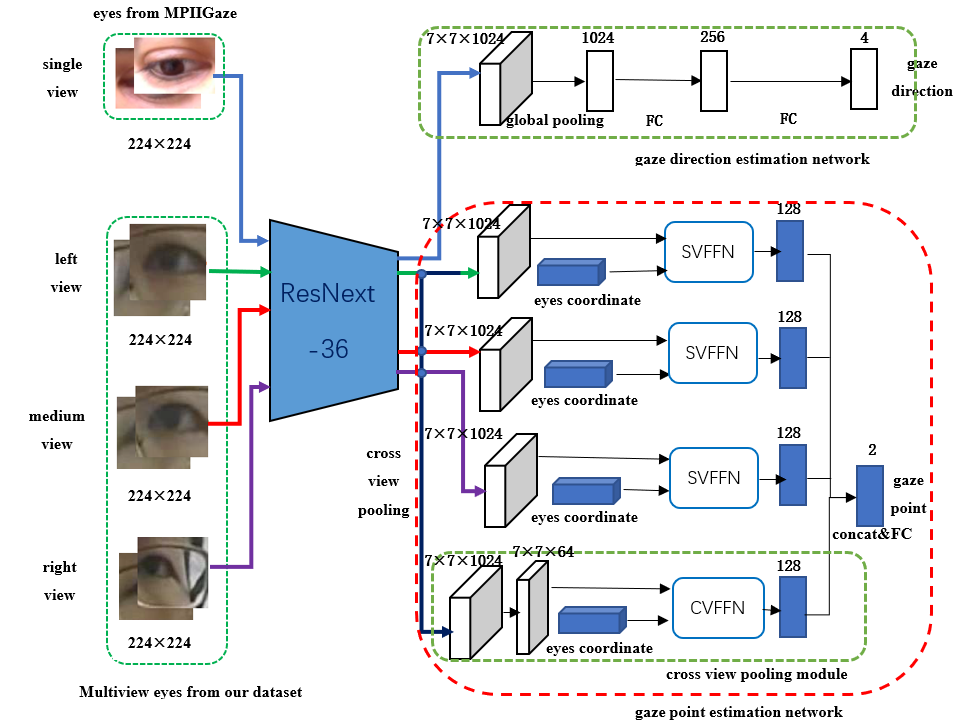}   %[]里可以指定影像大小
    \caption{Overall architecture of multiview multitask network for the 3-D gaze direction and 2-D gaze point estimations.}    % 图名
    \label{fig_electronic2}  % 用于内部引用的图名
\end{figure*}

% \begin{figure*} 
%     \centering
% 	\subfloat[a]{
% 	   \centering
%       \includegraphics[width=0.46\linewidth]{3.png}}
%     \label{1a}\hfill
% 	  \subfloat[b]{
% 	    \centering
%         \includegraphics[width=0.46\linewidth]{4.png}}
%     \label{1b}\\
% 	  \caption{(a), (b) Some examples from CIFAR-10 \cite{4}. The objects in     
%         single-label images are usually roughly aligned.(c),(d) However, the 
%         assumption of object alignment is not valid for multi-label
%         images. Also note the partial visibility and occlusion
%         between objects in the multi-label images.}
% 	  \label{fig1} 
% \end{figure*}

\begin{figure*} %使用的是强制位置，除非真的放不下，不然就是写在哪里图就放在哪里，不会乱动
	\centering  %图片全局居中
	\vspace{-0.35cm} %设置与上面正文的距离
	\subfigtopskip=2pt %设置子图与上面正文或别的内容的距离
	\subfigbottomskip=2pt %设置第二行子图与第一行子图的距离，即下面的头与上面的脚的距离
	\subfigcapskip=-5pt %设置子图与子标题之间的距离
	\subfigure[Single View Feature Fusion Network(SVFFN)]{
		\label{level.sub.1}
		\includegraphics[width=0.46\linewidth]{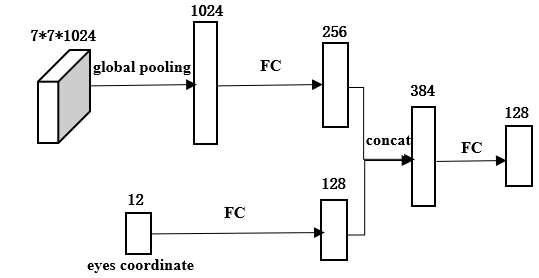}}
% 	\quad %默认情况下两个子图之间空的较少，使用这个命令加大宽度
	\subfigure[Cross View Feature Fusion Network(CVFFN)]{
		\label{level.sub.2}
		\includegraphics[width=0.46\linewidth]{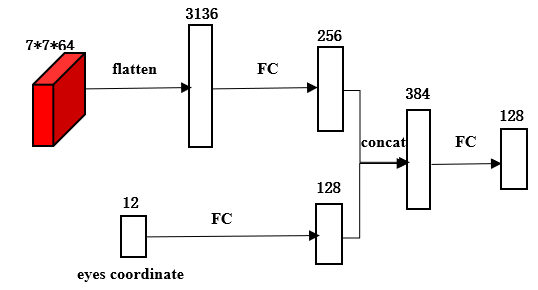}}
	\caption{(a), (b) are specific architecture of Single View Feature Fusion Network(SVFFN) and Cross View Feature Fusion Network(CVFFN) as show in Figure 2.}
	\label{level}
\end{figure*}

Input the eye image with the size of 224×224 including the gaze point and gaze direction into the local shared residual network ResNext-36. After feature extraction by ResNext-36, the output of each eye is 7×7×512 feature map after feature extraction, 7×7 represents the spatial resolution of the feature map, and 512 represents the dimension of the feature map. Then concatenate the feature maps of the two eyes in the same image, a the dimension is transformed to 1×1×1024, after passing through two fully connected layers, it becomes a 4-dimensional gaze direction; the gaze point estimation prediction module is similar to the gaze direction prediction module, which extracts features from images of 3 fields of view, and finally obtains 4 types of feature maps. Three types are obtained by extracting features of various field of view images separately with ResNext-36. The fourth type of feature map is obtained by cross fusion of the first three types. The scale of each type of feature map is 7×7×1024. Four types of feature maps construct cross-view pooling, and finally use full connection to connect the correspondinnd the scale of the concatenated feature maps becomes 7×7×1024. The gaze direction is estimated using the feature maps. After the feature maps are globally pooled,g features of the left, middle and right, cross-view pooling branches, and the output two-dimensional vector is the predicted gaze point

\subsection{Multi-Task learning loss function }
we mainly analyze and study the multi-view and multi-task eye gaze estimation algorithm proposed in this paper. For the eye gaze estimation network structure based on the convolutional neural network proposed in the first section, we need to design the cost functions of gaze direction and gaze point estimation respectively. Whether the two cost functions are set properly or not will directly affect the final gaze estimation network model. Performance. After mathematical theoretical derivation and experimental research and investigation, we proposed the loss function of gaze direction estimation and gaze point estimation. In order to make the task of gaze direction estimation and gaze point estimation can be completed in a network model at the same time, that is, using a local shared network in a network The model realizes the prediction of the gaze point and the gaze direction at the same time. We need to combine these two separate tasks into one. The combined loss function must complete the two tasks of gaze direction and gaze point estimation at the same time, so it is called multi-task gaze estimation . Finally, the parameters of the multi-view multi-task eye gaze estimation network based on the local network sharing mechanism are explained

\vspace{-0.25cm}
\noindent 
\textbf{Gaze direction estimation loss: } for the MPIIGaze dataset, the Cartesian coordinates of the three-dimensional vector are used to represent the gaze direction. However, in the Cartesian coordinate system, a small calculation error will cause the actual gaze direction to deviate greatly, so the difference between the real value and the prediction is directly minimized. The Euclidean distance between them will cause inaccurate gaze direction estimation. In view of the fact that spherical coordinates have greater advantages over Cartesian coordinates in direction measurement, in order to reduce the calculation error in the process of gaze direction estimation, we convert the three-dimensional Cartesian coordinates of the gaze direction into the corresponding three-dimensional spherical coordinates. Suppose the gaze direction vector in the three-dimensional Cartesian coordinate system corresponding to the gaze point in the i-th image is expressed as $g^{c}(I_{i})=(x,y,z)$, the corresponding gaze direction vector in the spherical coordinate system is expressed as $g^{s}(I_{i})=(\theta,\phi,r)$. Loss function is:
$$
{l}_{1}={\frac{1}{N}}{\sum_{i=1}^N(\left \| g^{s}_{l}(I_{i})-\hat{g}^{s}_{l}(I_{i})  \right \|^2+\left \| g^{s}_{r}(I_{i})-\hat{g}^{s}_{r}(I_{i})  \right \|^2)}$$
${l,r}$ denote left eye gaze estimation and right eye gaze estimation respectively.
As show Figure 1,we know that eye gaze point depend on eye position and gaze direction,${o}$ is eye position in three-dimensional coordinate system,${g}^{c}_{l},{g}^{c}_{r}$ are expressed as left eye gaze direction and right eye gaze direction respectively,${g}_{p}$ is two-dimensional gaze point,${v}$ is the distance vector from left eye to right eye，they satisfy the coplanar relationship,therefore,coplanar loss function is:
$${l}_{2}={\frac{1}{N}}{\sum_{i=1}^N\left \| g^{c}_{l}(I_{i})\times{g}^{c}_{r}(I_{i})\cdot{v}(I_{i})  \right \|^2}$$
$g^{c}_{l}(I_{i})\times{g}^{c}_{r}(I_{i})$represents the cross product of the gaze direction vectors of the left and right eyes,${g}^{c}_{r}(I_{i})\cdot{v}(I_{i})$represents the inner product of the gaze vector and the distance vector.

combine the loss function corresponding to the truth and prediction with the coplanar loss function to obtain the gaze direction estimation loss function,gaze direction estimation loss function is ${l}_{direc}$:
\begin{equation*}
\begin{aligned}
{l}_{direc}=&{\frac{1}{N}}{\sum_{i=1}^N(\left \| g^{s}_{l}(I_{i})-\hat{g}^{s}_{l}(I_{i})  \right \|^2+\left \| g^{s}_{r}(I_{i})-\hat{g}^{s}_{r}(I_{i})  \right \|^2)} \\
&+\lambda_{1}\cdot{\frac{1}{N}}{\sum_{i=1}^N\left \| g^{c}_{l}(I_{i})\times{g}^{c}_{r}(I_{i})\cdot{v}(I_{i})  \right \|^2}
\end{aligned}
\end{equation*}

where $\lambda_{1}$is expressed as the weight coefficient of the coplanar loss

The gaze point of the k-th image $I_{k}$ is represented  $p(I_{k}$ by the corresponding predicted estimated gaze point. Supposing the number of true gaze points of all images in the data set is $M$, the gaze point estimation loss function ${l}_{point}$ can be obtained, which uses the least square error and takes the target value Minimize the sum of squares of the difference from the estimated value
$${l}_{point}={\frac{1}{M}}{\sum_{i=1}^M\left \| p(I_{k}) - \hat{p}(I_{k})  \right \|^2}$$

We combine different view images captured by cameras of different views and propose the multi-view and multi-task (MTL) framework of this paper. The entire network architecture is shown in Figure 2. Specifically, we merge the features output from different sub-network views and merge them into a fully connected layer for gaze point estimation. In addition, we share the parameters of the sub-network from different perspectives to reduce the number of parameters. This article also introduces a cross-view pooling layer to aggregate characteristics from different views. Such a cross-view pooling layer is very useful when there is a lack of detailed information or noise in some views, for example, information caused by light reflection from glasses loss. Then, we connect the advanced cross-view pool function with the original single-view advanced function for gaze prediction. The multi-view multi-task human eye gaze estimation loss function proposed in this paper is as follows:
\begin{equation*}
\begin{aligned}
{l}_{muiltiview-multitask}={\frac{1}{N}}\sum_{i=1}^N(\left \| g^{s}_{l}(I_{i})-\hat{g}^{s}_{l}(I_{i})  \right \|^2\\+\left \| g^{s}_{r}(I_{i})-\hat{g}^{s}_{r}(I_{i})  \right \|^2)+\lambda_{1}{\frac{1}{N}}{\sum_{i=1}^N\left \| g^{c}_{l}(I_{i})\times{g}^{c}_{r}(I_{i})\cdot{v}(I_{i})  \right \|^2}\\+\lambda_{2}{\frac{1}{M}}{\sum_{k=1}^M\left \| p(I^L_{k},I^M_{k},I^R_{k}) - \hat{p}(I^L_{k},I^M_{k},I^R_{k})\right \|^2}+\lambda_{3}{\frac{1}{2}}\left \| w\right\|^2  
\end{aligned}
\end{equation*}
where $I^L_{k},I^M_{k},I^R_{k}$,represent the images taken by the left, center, and right cameras at the kth gaze point in the gaze point dataset respectively,$I_{i}$ represent the i-th image in the gaze direction dataset.

\subsection{training details}
The network model is implemented based on the Pytorch framework in this paper. The eye area image size in MPIIGaze data used for gaze direction estimation is 224×224. By adopting a local network sharing mechanism, it is necessary to maintain the input image using gaze point and gaze direction estimation. The scale is consistent. Therefore, the multi-view data set for gaze estimation collected in this paper needs to be scaled to 224×224 after the eye region is extracted.
In the multi-view multi-task gaze estimation network training, a deep supervision mechanism is added in the process of extracting features for each view. The SGD algorithm is used for optimization in the experiment here. The hyperparameter settings when training the network are as follows: Regarding the two data sets of multi-tasking, the batch size of the small sample input of MPIIGaze and our data set collected by the experiment is 128 and 32 respectively, which is a ratio of 4:1. The learning rate is set to $10^{-5}$, the momentum of the SGD optimizer with momentum is set to 0.9, the weight attenuation coefficient is, the number of alternations is 80, and the coefficient of the coplanar loss function in the training phase is variable $10^{-6}$, the purpose is to make the gaze direction the sum in the loss function is kept at the same magnitude, otherwise the party in the order of magnitude series will perform the entire optimization process, and the weight coefficient of the gaze point estimation module is $10^{-3}$.

\section{Experiment}
Due to the lack of multi-view data set and the insufficient data feature representation, this paper constructs a multi-view data set. Aiming at the data acquisition experiment for multi-view gaze point estimation, this paper uses 3 calibrated cameras, which are placed in the middle of the bottom of the display and about 15cm away from the middle of the left and right ends, and shoot people at the same time under natural conditions. Obtain the three-field image, perform face detection on the collected three-field image, extract the eye area feature from the facial feature through a certain algorithm, and use the corresponding eye area image as the data set for gaze point estimation in this article. Ours data set also contains eye position information and real gaze points. The data set needs further data preprocessing before importing it into the network. Data set for gaze direction estimation This article uses the published MPIIGaze data set, which contains single-view eye images, real gaze direction, and head posture information.
The preprocessed gaze point estimation dataset ours and gaze direction estimation dataset MPIIGaze are both input to the local sharing network ResNext-36 with an image size of 224×224, MPIIGaze is input as a single image, and the corresponding ours is in the left, middle and right Three view image input; ResNext-36 network uses convolutional layer, pooling layer, etc. to perform feature extraction and dimensionality reduction, and the output is all 7×7×1024 feature maps.
Pass the output corresponding feature map to the gaze estimation network and the gaze point estimation network, propagate forward to the loss layer, and bring the predicted value and the true value into the loss function to calculate the loss value. When the loss value is greater than the set threshold, the model continues Training, back propagation update gradient from the loss layer, update the network parameters from deep to shallow, and then forward propagation until the loss value is less than a certain threshold, the model training is completed.The gaze estimation result of the network model is the gaze point and gaze direction, so it is necessary to evaluate the estimated gaze point and gaze direction. Because a single sample has a certain contingency, this article uses the sample mean in the evaluation index.

gaze point evaluation method:
$${E}_p={\frac{1}{M}}{\sum_{i=1}^M\left \| {p_i} - \hat{p_i}  \right \|}$$
${p_i}$ represents the true gaze point of the i-th picture, represents the gaze point prediction of the i-th picture, M represents the total number of pictures, and ${E}_p$ represents the average distance error between the real gaze point and the predicted gaze point,unit is centimeters.   
gaze direction evaluation method:
$${E}_d={\frac{1}{N}}{\sum_{i=1}^N\arccos\frac{{g_i}\cdot{\hat{g_i}}}{\left \| {g_i}\right\| \left \| \hat {g_i}  \right \|}}$$
${g_i}$ represents the real gaze angle of the i-th picture, $\hat{g_i}$ represents the gaze angle predicted by the network of the i-th picture, and ${E}_d$ represents the average angle deviation between the predicted angle and the real angle,unit is angles.

In the following comparison experiment, the gaze point is estimated to be measured as a two-dimensional coordinate, and the gaze direction is measured as a three-dimensional direction vector. Before the performance comparison, we compare the experimental data sets in this article. The three data sets are Ours, UT Multiview, and MPIIGaze data sets collected in this article. The first two data sets are multi-views for gaze point estimation. The MPIIGaze data set is a single view for gaze direction estimation. In the experiment, it is divided into training set and test set according to the ratio of 8:2. Table 1 is a comparison of the three data sets.

% table first

\setlength{\tabcolsep}{0.5mm}{
\begin{table}
\centering
\caption{Comparison of three data sets}%写标题
\label{tab:1}
\begin{tabular}{cccccc}
    
	\toprule
		dataset&person&camera&light&picture&point or angle \\
	\midrule
		UT Multiview&50&8&y&64000&p\\
		
		MPIIGaze&15&1&n&213659&a\\
		Ours&137&3&y&233796&p\\
	\bottomrule
\end{tabular}
\end{table}
}

\subsection{Multi-task and single-task performance}
In order to quantitatively measure the effect of the multi-task learning mechanism, first we use the experimentally collected multi-view ours data set and UT Multiview data set to build a network model for gaze point estimation only. The network model is similar to Figure 3.1.3 in Chapter 3. a) The network models shown are consistent, and only the gaze direction estimation branch module is eliminated. Next, use the MPIIGaze dataset to construct a network model that only estimates the gaze direction using the same method. The two single-task networks are trained and learned separately. Finally, the multi-task learning mechanism is used to train the network model to estimate the gaze point and gaze direction at the same time. The gaze point and gaze direction evaluation indicators are used to calculate the single-task gaze point, the single-task gaze direction, and the multi-task gaze point and the gaze direction estimated average deviation, which are used as performance indicators for experimental comparison. Single task and multitask in the experimental collection of ours data set, UT Multiview data set gaze point performance comparison, the performance comparison in the single view MPIIGaze data set is shown in Table 2.

%table second
\begin{table}[!htbp]
  \centering
  \caption{Multi-task and single-task staring estimation performance comparison}%写标题
  \label{tab:2}
  \begin{tabular}{|c|c|c|c|c|c|c|}
  \hline
  \multicolumn{1}{|c|}{\multirow{2}*{\diagbox[innerwidth=1.6cm]{tasks}{datasets}}}& \multicolumn{3}{c|}{MPIIGaze | UT} &\multicolumn{3}{c|}{MPIIGaze | Ours}\\
  \cline{2-7}
  \multirow{2}*{}&l-eye&r-eye&point&l-eye&r-eye&point\\
  \hline
  \multirow{3}*{}single-task&$4.83^\circ$&$4.84^\circ$&$4.57cm$&$4.83^\circ$&$4.84^\circ$&$5.26cm$\\
  \hline
  \multirow{4}*{}multi-task&$4.7^\circ$&$4.7^\circ$&$3.9cm$&$4.6^\circ$&$4.6^\circ$&$4.61cm$\\
  \hline
  \end{tabular}
 \end{table}
 
Analysis of Table 2 shows that the performance of the multi-task learning mechanism is superior to the single-task mechanism in terms of gaze point and gaze direction estimation. As explained before, the gaze point is closely related to the gaze direction. The multitasking mechanism can supervise the network to optimize the parameters from two different domains, improve the performance of the gaze point estimation task, and at the same time help the gaze direction estimation task to optimize the parameters. Optimize the gaze point estimation task. Although our data set does not have both the real value of the gaze direction and the real value of the gaze point, forcing different data sets to share a network is helpful for parameter optimization, while reducing network overhead, shortening training time, and improving the performance of the two tasks. performance.

\subsection{Multi-view and single-view performance}
In order to verify the validity of the multi-view data, this article also compares the single-view and multi-view experiments between the ours data set and the UT Multiview data set collected in the experiment. The specific method is that the left, center, and right are used as a single view and combined with the left, center, and right. The multi-view for comparison. All experiments are performed on a multi-task learning framework, and the MPIIGaze dataset is used to estimate the gaze direction. Gaze point estimation performance of left view, middle view, right view and multi-view under different models
\\
%table third
\setlength{\tabcolsep}{0.5mm}{
\begin{table}
\centering
\caption{Performance comparison between multi-view and single-view}%写标题
\label{tab:3}
\begin{tabular}{ccccc}
	\toprule
		method&l-view&m-view&r-view&mul-view\\
	\midrule
		MPIIGaze+UT+l+d&$4.76^\circ$&$4.75^\circ$&$4.79^\circ$&$4.73^\circ$ \\
		MPIIGaze+UT+r+d&$4.77^\circ$&$4.76^\circ$&$4.77^\circ$&$4.72^\circ$ \\
		UT+p&$4.82cm$&$4.76cm$&$4.83cm$&$3.89cm$ \\
	\midrule	
	    MPIIGaze+Ours+l+d&$4.60^\circ$&$4.63^\circ$&$4.64^\circ$&$4.53^\circ$ \\
		MPIIGaze+Ours+r+d&$4.63^\circ$&$4.64^\circ$&$4.70^\circ$&$4.53^\circ$ \\
		Ours+p&$5.49cm$&$5.56cm$&$5.60cm$&$4.60cm$\\
	\midrule	
		Ours+no+glass&$5.00cm$&$4.96cm$&$4.90cm$&$4.13cm$\\
        Ours+glass&$5.73cm$&$5.86cm$&$5.92cm$&$4.81cm$ \\

	\bottomrule
\end{tabular}
\end{table}
}

the multi-view image improves the gaze point by a greater extent, but it improves the gaze direction less. Because the data sets for gaze direction estimation are all MPIIGaze data sets, the effect of improving gaze direction estimation is very limited. Since the estimation of the gaze point depends on the gaze direction and the eye position, this implies that the multi-view image helps to determine the accurate eye position information. At the same time, in order to verify the influence of the method in this paper on the glasses with and without glasses, this paper divides the data into two groups with glasses and without glasses for experiments. From Table 3, it can be seen that images without glasses are more conducive to gaze estimation. Because the images taken while wearing glasses are easily affected by occlusion, specular reflection, and color. Introduced the multi-view data experiment and found that the lifting effect of wearing glasses is better than that of images without eyes, which proves the importance of the multi-view structure in reducing the occlusion and noise of the human eye image.

\subsection{Performance based on deep learning methods}
Since the method in this article is based on deep learning, three classic gaze estimation networks based on deep learning are selected for comparison, which are multi-modal convolutional neural network [36], eye tracking network [5], spatial weighted convolution Neural networks [12], these networks are single-view gaze estimation. In order to be able to use multi-view input for gaze point estimation, this paper uses full connection to concatenate the features from all view layers, and make a series of changes based on this type of basic network. The resulting network models are: multi-view multi-modal convolutional neural Network, multi-view eye tracking network, multi-view spatial weighted convolutional neural network.
At the same time, in order to verify the effectiveness of the local shared network in this article, the experiment also changed the basic network of the multi-modal convolutional neural network [36], eye tracking network [5], and spatial weighted convolutional neural network [12] to this article Partially shared network ResNext-36, other structures remain unchanged.
The above two indicators are compared with the above gaze estimation network model. Aiming at the comparison of the effect of gaze point estimation, in order to ensure the reliability of the results, we compare our own experimental collection of ours and UT Multiview two multi-view data sets. The comparison of gaze direction estimation effects are all on the MPIIGaze single-view data set. The experimental results of the average angle error of the gaze direction estimation of the network model in this paper on the MPIIGaze dataset are shown in Table 4 (unit: degree). The experimental results of the gaze point estimation average distance deviation of the network model in this paper on ours data set and UT Multiview data set are shown in Table 5.

% table four
\setlength{\tabcolsep}{0.5mm}{
\begin{table}
\centering
\caption{Comparison of average angle error of gaze direction estimation on MPIIGaze dataset}%写标题
\label{tab:4}
\begin{tabular}{ccc}
	\toprule
		model&left eye&right eye \\
	\midrule
		Muiti-modal[36]&$6.62^\circ$&$6.57^\circ$\\
	Spatial weight CNN[12]&$6.02^\circ$&$6.04^\circ$\\
	Eye tracking[5]&$6.05^\circ$&$6.06^\circ$\\
	Eye tracking based on ResNet-34[5]&$5.29^\circ$&$5.31^\circ$\\
	\midrule
	    Our model+ UT dataset&$4.80^\circ$&$4.80^\circ$\\
	     Our model+ Our dataset&$4.70^\circ$&$4.70^\circ$\\
	\bottomrule
\end{tabular}
\end{table}
}

% table five
\setlength{\tabcolsep}{0.5mm}{
\begin{table}
\centering
\caption{Comparison of the average error of gaze point estimation on ours and UT Multiview data sets}%写标题
\label{tab:5}
\begin{tabular}{ccc}
	\toprule
		model&Ours&UT \\
	\midrule
		Multi-view Muiti-modal[36]&$5.41^\circ$&$6.33^\circ$\\
		Multi-view Muiti-modal[12]&$5.15^\circ$&$5.82^\circ$\\
		Multi-view Eye tracking[5]&$5.01^\circ$&$5.80^\circ$\\
	Multi-view Eye tracking based ResNet[5]&$4.38^\circ$&$5.08^\circ$\\
	\midrule
	    Multi-view Muiti-modal+ResneXt36&$4.71^\circ$&$5.32^\circ$\\
	    Multi-view Single-modal+ResneXt36&$4.58^\circ$&$5.19^\circ$\\
    \midrule
	    Ours&$3.9^\circ$&$4.8^\circ$\\
	\bottomrule
\end{tabular}
\end{table}
}

analyzing Tables 4 and 5, we can see that the average error of the method in this paper is lower than that of the existing mainstream methods in terms of gaze point and gaze direction estimation, showing a greater performance improvement. The gaze point estimation results of the multi-view multi-modal convolutional neural network and the local shared network ResNext-36 in Table 5 also verify that the local shared network ResNext-36 in this paper has more advantages in feature extraction.

\section{Acknowledgements}
The authors wish to thank HuaZhong university of science and technology for providing computationaland storage resources.Of course thanks my supervisor by Prof. Daiming Qu to careful guidance,the code would open in https://github.com/AderonHuang.

%------------------------------------------------------------------------
%第一种方法加入参考文献
% {\small
% \bibliographystyle{ieee}
% \bibliography{egbib}
% }

% 第二种方法加入参考文献

\end{document}